# Adaptive Non-linear Filtering Technique for Image Restoration

S. K. Satpathy, S. Panda, K. K. Nagwanshi, S. K. Nayak and C. Ardil

*Abstract*—Removing noise from the any processed images is very important. Noise should be removed in such a way that important information of image should be preserved. A decision-based nonlinear algorithm for elimination of band lines, drop lines, mark, band lost and impulses in images is presented in this paper. The algorithm performs two simultaneous operations, namely, detection of corrupted pixels and evaluation of new pixels for replacing the corrupted pixels. Removal of these artifacts is achieved without damaging edges and details. However, the restricted window size renders median operation less effective whenever noise is excessive in that case the proposed algorithm automatically switches to mean filtering. The performance of the algorithm is analyzed in terms of Mean Square Error [MSE], Peak-Signal-to-Noise Ratio [PSNR], Signal-to-Noise Ratio Improved [SNRI], Percentage Of Noise Attenuated [PONA], and Percentage Of Spoiled Pixels [POSP]. This is compared with standard algorithms already in use and improved performance of the proposed algorithm is presented. The advantage of the proposed algorithm is that a single algorithm can replace several independent algorithms which are required for removal of different artifacts.

*Keywords*—Filtering, Decision Based Algorithm, noise, image restoration.

## I. INTRODUCTION

THE idea of image restoration is to balance for or unwrap defects which corrupts an image. Degradation comes in many forms such as motion blur, noise, and camera misfocus. In cases like motion blur, it is possible to come up with a very good estimate of the actual blurring function and undo the blur to restore the original image. In cases where the image is corrupted by noise, the best we may hope is to compensate for the degradation it caused. In this paper, it will introduce and implement several of the methods used in the image processing world to restore images. There are two different domains under which one can go for image restoration: (i) Spatial Domain and (ii) Frequency Domain. This paper is fully concentrated on frequency domain method. It will discuss some existing methods along with proposed approach. It is well known that linear filters are not quite effective in the presence of non-Gaussian noise. In the last decade, it has been shown that nonlinear digital filters can overcome some of the limitations of linear digital filters [1]. Median filters are a class of nonlinear filters and have produced good results where linear filters generally fail [2]. Median filters are known to remove impulse noise and preserve edges. There are a wide variety of median filters in the literature. In remote sensing, artifacts such as strip lines, drop lines, blotches, band missing occur along with impulse noise. Standard median filters reported in the literature do not address these artifacts. Strip lines are caused by unequal responses of elements of a detector array to the same amount of incoming electromagnetic energy [3]. This phenomenon causes heterogeneity in overall brightness of adjacent lines. Drop line [3] occurs when a detector does not work properly for a short period. Impulse noise appears when disturbing microwave energies are present or the sensor/detector is degraded. Band missing [3] is a serious problem and is caused by corruption of two or more drop/strip lines continuously. For removal of these artifacts, generally separate methods are employed. Strip lines and drop lines are considered as line scratches by Silva and Corte-Real [4] for image sequences. According to him, a positive type film suffers from bright scratches and negative film suffers from dark scratches. Milady has considered only the dark scratches; if bright scratches exist he inverted them and used the same algorithm. Silva and Corte-Real [4] gives a solution for removing the blotches and line scratches in images. He has considered only vertical lines (which are narrow) and the blotches as impulsive with constant intensity having irregular shapes. Kokaram [5] has given a method for removal of scratches and restoration of missing data in the image sequences based on temporal filtering. Additionally, impulse noise is a standard type of degradation in remotely sensed images. This paper considers application of median-based algorithms for removal of impulses, strip lines, drop lines, band missing, and blotches while preserving edges. It has been shown recently that an adaptive length algorithm provides a better solution for removal of impulse noise with better edge and fine detail preservation. Several adaptive algorithms [6–9] are available for removal of impulse noises. However, none of these algorithms addressed the problem of strip lines, drop lines, blotches, and band missing in images.

S.K. Satpathy is with the Department of Computer Science & Engineering, Rungta College of Engineering & Technology, Bhilai CG 490023 India (Phone: 922-935-5518; fax: 788-228-6480; e-mail: sks_sarita@yahoo.com).

S. Panda is with the National Institute of Science and Technology (NIST) Berhampur, OR 760007 India (e-mail: panda_siddhartha@rediffmail.com).

K.K. Nagwanshi is with the Department of Computer Science & Engineering, Rungta College of Engineering & Technology, Bhilai CG 490023 India (e-mail: kapilkn@gmail.com).

S.K. Nayak is with Department of ECE, Aditya Institute of Technology, Tekkali, Srikakulam Dt. AP. 532201(e-mail:sknayakbu@rediffmail.com)

C. Ardil is with National Academy of Aviation, AZ1045, Baku, Azerbaijan, Bina, 25th km, NAA (e-mail: cemalardil@gmail.com)

The objective of this paper is to propose an adaptive length median/mean algorithm that can simultaneously remove impulses, strip lines, drop lines, band missing, and blotches while preserving edges. The advantage of the proposed algorithm is that a single algorithm with improved performance can replace several independent algorithms required for removal of different artifacts. Blotches are impulsive-type degradations randomly distributed with irregular shapes of approximately constant intensity. These artifacts last for one frame. In the degraded regions, there is no correlation between successive frames. Blotches are originated by dust, warping of the substrate or emulsion, mould, dirt, or other unknown causes. Blotches in film sequences can be either bright or dark spots. If the blotch is formed on the positive print of the film, then the result will be a bright spot, however if it is formed on the negative print, then in the positive copy, we will see a dark spot.

The paper has been organized in the following manner; section II describes image noise model, section III proposes problem formulation and solution methodology, proposed algorithm is illustrated in section IV, followed by section V which gives implementation details subsequently section VI describes the result and discussion, section VII gives concluding remarks and further works and finally section VIII incorporates all the references been made for completion of this work.

## II. IMAGE NOISE MODEL

An image usually contains departures from the ideal signal that would be produced by general model of the signal production process. Such departures are referred to as noise. Noise arises as a result of un-modeled processes available on in the production and capture of the actual signal. It is not part of the ideal signal and may be caused by a wide range of sources, e.g. variations in the detector sensitivity, environmental variations, the discrete nature of radiation, transmission or quantization errors, etc. It is also possible to treat irrelevant scene details as if they are image noise (e.g. surface reflectance textures). The characteristics of noise depend on its source, as does the operator which best reduces its effects.

Many image processing tools deal with contain operators to artificially add noise to an image. Deliberately corrupting an image with noise allows us to test the resistance of an image processing operator to noise and assess the performance of various noise filters. Noise can generally be grouped into two classes: (i) independent noise, and (ii) noise which is dependent on the image data. *Image independent noise* can often be described by an additive noise model, where the traced image $f(x,y)$ is the sum of the original image $s(x,y)$ and the noise $n(x,y)$: i.e. $f(x,y)=s(x,y)+n(x,y)$.

In several cases, additive noise is equally distributed over the frequency domain (i.e. white noise), while an image contains generally low frequency information. Hence, the noise is dominant for high frequencies and its effects can be reduced using some kind of low-pass filter. This can be done either with a frequency filter or with a spatial filter. (Often a spatial filter is preferable, as it is computationally less expensive than a frequency filter.) In the second case of data-dependent noise (e.g. arising when monochromatic radiation is scattered from a surface whose roughness is of the order of a wavelength, causing wave interference which results in image speckle), it is possible to model noise with a multiplicative, or non-linear model. These models are mathematically more complicated. Hence, the noise can be assumed as data independent.

*Detector Noise* is a kind of noise which occurs in all traced images to a certain extent is detector noise. This kind of noise is due to the discrete nature of radiation, i.e. the fact that each imaging system is recording an image by counting photons. Allowing some assumptions (which are valid for many applications) this noise can be modeled with an independent, additive model, where the noise $n(x,y)$ has a zero-mean Gaussian distribution described by its standard deviation ($\sigma$), or variance. (The 1-D Gaussian distribution has the form shown in Figure 1.) This means that each pixel in the noisy image is the sum of the true pixel value and a random, Gaussian distributed noise value.

*Salt and Pepper Noise* is another common form of noise is data drop-out noise (commonly referred to as intensity spikes, speckle). Here, the noise causes errors in the data transmission. The corrupted pixels are either set to the maximum value (which looks like snow in the image) or have single bits flipped over. In some cases, single pixels are set alternatively to zero or to the maximum value, giving the image a *salt and pepper* like appearance. Unaffected pixels always remain unchanged. The noise is usually quantified by the percentage of pixels which are corrupted.

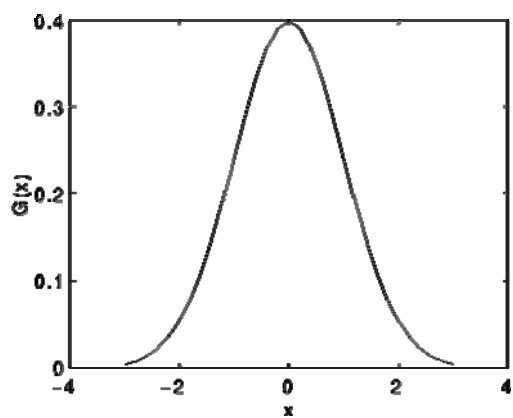

Fig. 1 1-D Gaussian distribution (mean=0 standard deviation=1)

The block diagram (fig. 2) for general degradation model is where $f(x,y)$ is the corrupted image obtained by passing the original image $s(x,y)$ through a low pass filter (blurring function) $b(x,y)$ and adding noise $n(x,y)$ to it.

It is difficult to propose a general mathematical model for the effect of the abrasion of the film causing the scratches due to the high number of variables that are involved in the process. However, it is possible to make some physical and geometrical considerations regarding the brightness, thickness, and vertical extent of the line. Line scratches can be characterized as follows: (i) they present a considerable higher or lower luminance than their neighborhoods; (ii) they tend to extend over most of the vertical length of the image frame and are not curved; and (iii) they are quite narrow, with widths no larger than 10 pixels for video images. These features can be used to define a model. The degraded image model considered is

$$a(x,y)=I(x,y)(1-b(x,y))+b(x,y)c(x,y), \quad (1)$$

where $I(x,y)$ is the pixel intensity of the uncorrupted signal, $b(x,y)$ is a detection variable which is set to 1 whenever pixels are corrupted and 0 otherwise, $c(x,y)$ is the observed intensity in the corrupted region. This model is applied in this work to images degraded by impulses, strip lines, drop lines, band missing, and blotches.

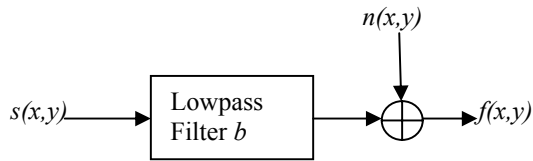

Fig 2 A general degradation model

If $b(x,y)=0$ then

$$a(x,y)=I(x,y)(1-0)+0.c(x,y)=I(x,y), \quad (2)$$

where $I(x,y)$ is the original pixel value (uncorrupted pixel).

If $b(x,y)=0$ then

$$a(x,y)=I(x,y)(1-1)+1.c(x,y)=c(x,y), \quad (3)$$

where $c(x,y)$ is the original pixel value (uncorrupted pixel).

Assume that each pixel at $(x,y)$ is corrupted by an impulse with probability p independent of whether other pixels are corrupted or not. For images corrupted by a negative or positive impulse, the impulse corrupted pixel $e(x,y)$ takes on the minimum pixel value $s_{min}$ with probability $p$, or $s(x,y)$ the maximum pixel value $s_{max}$ with probability $1-p$. The image corrupted by blotches or scratches (impulsive) can be now modeled as

$$c(x,y)=e(x,y) \quad \text{with } p$$
$$s(x,y) \text{ with } 1-p \quad (4)$$

This, in fact, is the model that describes impulse noise in the literature. However, the existing impulse filtering algorithms do not effectively remove blotches and scratches. In Section 3, an adaptive length median/mean filter algorithm is developed that removes blotches, scratches effectively along with impulse noise.

III. PROBLEM FORMULATION AND SOLUTION METHODOLOGY

Median filter is a nonlinear filter, which preserves edges while effectively removing impulse noise. Median operations are performed by row sorting, column sorting, and diagonal sorting in images [10]. General median filters often exhibit blurring for large window sizes, or insufficient noise suppression for small window sizes. Adaptive length median filter overcomes these limitations of general median filters. Lin and Willson [6] proposed an adaptive window length median filter algorithm which can achieve a high degree of noise suppression and still preserve image sharpness; however, the algorithm performs poorly for mixed impulse noise consisting of positive and negative impulses. Lin's algorithm is modified by Hwang and Haddad [7]. Huang's algorithm takes into account both positive and negative impulses for simultaneous removal; but it acts poorly on the strip lines, drop lines, and blotches.

Unlike these adaptive algorithms based on edge detection [6,7], the proposed algorithm is based on artifacts detection. The positive and negative impulses are removed separately. In contrast to general adaptive length median filters, the window size is restricted to a maximum of to minimize blurring. Restriction of window size renders the median operation less effective whenever noise is excessive (the output of the median filter may turn out to be a noisy pixel). In this situation, the algorithm switches to compute the average of uncorrupted pixels in the window (the probability of getting the noisy pixel as filtered output is lower because the averaging takes only uncorrupted pixels into account). The proposed algorithm removes the strip lines, drop lines, blotches along with impulses even at higher noise densities.

Image enhancement in the frequency domain is straightforward. We simply compute the Fourier transform of the image to be enhanced, multiply the result by a filter (rather than convolve in the spatial domain), and take the inverse transform to produce the enhanced image. The idea of blurring an image by reducing its high frequency components or sharpening an image by increasing the magnitude of its high frequency components is intuitively easy to understand. However, computationally, it is often more efficient to implement these operations as convolutions by small spatial filters in the spatial domain. Understanding frequency domain concepts is important, and leads to enhancement techniques that might not have been thought of by restricting attention to the spatial domain.

*A. Filtering*

Low pass filtering involves the elimination of the high frequency components in the image. It results in blurring of the image (and thus a reduction in sharp transitions associated with noise). An ideal low pass filter would retain all the low frequency components, and eliminate all the high frequency components. However, ideal filters suffer from two problems: *blurring* and *ringing*. These problems are caused by the shape of the associated spatial domain filter, which has a large number of undulations. Smoother transitions in the frequency domain filter, such as the Butterworth filter, achieve much better results.

*B. Homomorphic filtering*

Images normally consist of light reflected from objects. The basic nature of the image $G(x,y)$ may be characterized by two components: (1) the amount of source light incident on the scene being viewed, and (2) the amount of light reflected by the objects in the scene. These portions of light are called the *illumination* and *reflectance* components, and are denoted $i(x,y)$ and $r(x,y)$ respectively. The functions $i$ and $r$ combine multiplicatively to give the image function $f(x,y) = i(x,y)*r(x,y)$, Where $0 < i(x,y) < \infty$ and $0 < r(x,y) < 1$. We cannot easily use the above product to operate separately on the frequency components of illumination and reflection because the Fourier transform of the product of two functions is not separable; that is $G(f(x,y)) \neq G(i(x,y))*G(r(x,y))$ Suppose, however, that we define $z(x,y)$ *is equal to* $\ln f(x,y)$ or we may also say that it is equal to $\ln i(x,y) + \ln r(x,y)$ Then $G(z(x,y))=G(\ln f(x,y))$ one can describe it as sum of $G(\ln i(x,y)$ and $\ln r(x,y))$
or

$$Z(\omega,v)=I(\omega,v)+R(\omega,v), \quad (5)$$

where $Z$, $I$ and $R$ are the Fourier transforms of $z$, $\ln i$ and $\ln r$ respectively. The function $Z$ represents the Fourier transform of the *sum* of two images: a low frequency illumination image and a high frequency reflectance image.

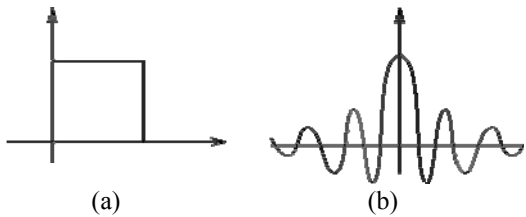

(a)        (b)

Fig. 3 Transfer function for an ideal low pass filter (a) Frequency domain components (b) spatial domain components.

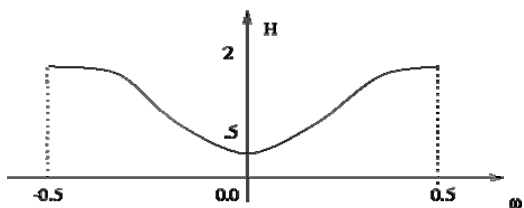

Fig. 4 Transfer function for homomorphic filtering.

If we now apply a filter with a transfer function that suppresses low frequency components and enhances high frequency components, then we can suppress the illumination component and enhance the reflectance component. Thus

$S(\omega,v)=H(\omega,v)*Z(\omega,v)$

$= H(\omega,v)*I(\omega,v)+H(\omega,v)*R(\omega,v)$ (6)

where $S$ is the Fourier transform of the result. In the spatial domain

$s(x,y)=G^{-1}(S(\omega,v))$

$=G^{-1}(H(\omega,v)*I(\omega,v))+G^{-1}(H(\omega,v)*R(\omega,v))$ (7)

By letting

$i'(x,y)= G^{-1}(H(\omega,v)*I(\omega,v))$

and

$r'(x,y)=G^{-1}(H(\omega,v)*R(\omega,v))$

we get

$s(x,y) = i'(x,y) + r'(x,y).$ (8)

Finally, as $z$ was obtained by taking the logarithm of the original image $G$, the inverse yields the desired enhanced image $\hat{G}$ that is

$\hat{G}(x,y)=exp[s(x,y)]$

$= exp[i'(x,y)]*exp[r'(x,y)]$

$=i_0(x,y)*r_0(x,y)$ (9)

Thus, the process of homomorphic filtering can be summarized by the following diagram:

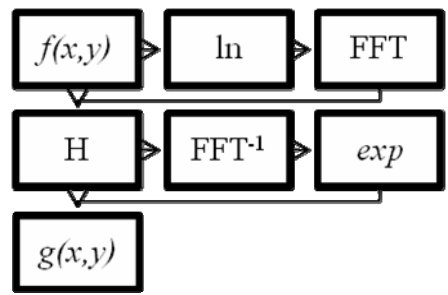

Fig. 5 The process of homomorphic filtering

### IV. PROPOSED ALGORITHM

The algorithm consists of two operations: first is the detection of degraded pixels, and the second operation is the replacement of faulty pixels with the estimated values.

Let the pixel be represented as $P(i,j)$ and the number of corrupted pixels in the window $W(i,j)$ be n Let $P_{max}=255$ and $P_{min}=0$ be the corrupted pixel values and $P(i,j) \neq 0$ or 255 represent uncorrupted pixels.

Case 1: Consider window size 3X3 with typical values of pixels shown as an array below. If $P(i,j) \neq 0$ or 255, then the pixels are unaltered. For the array shown, there are no corrupted pixels in the array; therefore, the pixels are unaltered.(10)

$$\begin{vmatrix} 111 & 214 & 106 \\ 236 & 167 & 214 \\ 123 & 223 & 83 \end{vmatrix} \quad (10)$$

Case 2: If the number of corrupted pixels n in the window $W(i,j) \leq 4$, that is, $n \leq 4$, then two-dimensional window of size 3X3 is selected and median operation is performed by column sorting, row sorting, and diagonal sorting. The corrupted $P(i,j)$ is replaced by the median value. (11)

| Corrupted matrix | | | Row sorting → | | |
|---|---|---|---|---|---|
| 255 | 214 | 123 | 123 | 214 | 255 |
| 0 | 255 | 214 | 0 | 214 | 255 |
| 123 | 234 | 0 | 0 | 123 | 234 |

| ↓Column sorting | | | Diagonal sorting↓ → | | | (11) |
|---|---|---|---|---|---|---|
| 0 | 123 | 234 | 0 | 123 | 123 | |
| 0 | 214 | 255 | 0 | 214 | 255 | |
| 123 | 214 | 255 | 234 | 214 | 255 | |

| Corrupted matrix | | | | | Row sorting → | | | | |
|---|---|---|---|---|---|---|---|---|---|
| 123 | 0 | 156 | 255 | 234 | 0 | 123 | 156 | 234 | 255 |
| 255 | 0 | 214 | 97 | 0 | 0 | 0 | 97 | 214 | 255 |
| 0 | 234 | 255 | 133 | 191 | 0 | 131 | 191 | 234 | 255 |
| 199 | 255 | 234 | 255 | 0 | 0 | 199 | 234 | 255 | 255 |
| 255 | 167 | 210 | 198 | 178 | 167 | 178 | 198 | 210 | 255 |

(12)

| Column sorting ↓ | | | | | Diagonal sorting | | | | |
|---|---|---|---|---|---|---|---|---|---|
| 0 | 0 | 156 | 97 | 0 | 0 | 0 | 98 | 210 | 167 |
| 123 | 0 | 210 | 133 | 0 | 0 | 123 | 156 | 178 | 255 |
| 199 | 167 | 214 | 198 | 178 | 0 | 133 | 191 | 234 | 255 |
| 255 | 234 | 234 | 255 | 234 | 0 | 214 | 198 | 234 | 255 |
| 255 | 255 | 255 | 255 | 255 | 255 | 199 | 234 | 255 | 255 |

Case 3: If the number of corrupted pixels n in the window $W(i,j)$ is between 5 and 12, that is, $5 \leq n \leq 12$, then perform 5X5 median filtering and replace the corrupted values by the median value.

Case 4: (i) If the number of corrupted pixels $n$ in the window $W(i,j)$ is greater than 13, (a typical case is shown as an array below) increasing the window size may lead to blurring; choose 3X3 median filtering. On median filtering with smaller window sizes, the output may happen to be noise pixels whenever the noise is excessive. In this case, find the average of uncorrupted pixels in the window and replace the corrupted value by the average value. The average of the pixel value in the window is taken instead of median value, if the number of uncorrupted pixels in the window is even (it is convenient to define median for odd number of pixels).

| (133 and 123 are the uncorrupted pixels) | | | | | 255 | 123 | 255 |
|---|---|---|---|---|---|---|---|
| | | | | | 255 | **255** | 133 |
| **123** | 0 | 0 | 255 | 0 | 0 | 255 | 0 |
| 255 | 255 | 123 | 255 | 0 | ↓ | | |
| 0 | 255 | 255 | **133** | 145 | 255 | 123 | 255 |
| 199 | 0 | 255 | 0 | 255 | 255 | **128** | 133 |
| 255 | 167 | 0 | 198 | 178 | 0 | 255 | 0 |

(13)
(133+123)/2=128

(ii) If all the pixels in 3X3 windows are corrupted (a typical case is shown as an array below), then perform 5X5 median filtering. On median filtering, the output may happen to be noise pixels as in Case 4. Find the average of uncorrupted pixels in the window and replace the corrupted value by the average value.

To assess the performance of the proposed filters for removal of impulse noise and to evaluate their comparative performance, different standard performance indices have been used in the thesis. These are defined as follows:

**Peak Signal to Noise Ratio (PSNR):** It is measured in decibel (dB) and for gray scale image it is defined as:

$$\text{PSNR (dB)} = 10\log_{10}\left[\frac{\sum_i\sum_j 255^2}{\sum_i\sum_j (S_{i,j}-\hat{S}_{i,j})^2}\right] \quad (14)$$

Where $S_{i,j}$ and $\hat{S}_{i,j}$ are the original and restored image pixels respectively. The higher the PSNR in the restored image, the better is its quality.

**Signal to Noise Ratio Improvement (SNRI):** SNRI in dB is defined as the difference between the Signal to Noise Ratio (SNR) of the restored image in dB and SNR of restored image in dB i.e.

SNRI (dB) = SNR of restored image in dB - SNR of noisy image in dB (15)

Where,
SNR of restored image dB=

$$10\log_{10}\left[\frac{\sum_i\sum_j S_{i,j}^2}{\sum_i\sum_j (S_{i,j}-\hat{S}_{i,j})^2}\right] \quad (16)$$

SNR of Noisy image in dB =

$$10\log_{10}\left[\frac{\sum_i\sum_j S_{i,j}^2}{\sum_i\sum_j (S_{i,j}-X_{i,j})^2}\right] \quad (17)$$

Where, $X_{i,j}$ is Noisy image pixel

The higher value of SNRI reflects the better visual and restoration performance.

**Percentage of Spoiled Pixels (POSP):** It may be defined as the number of unaffected original pixels replaced with a different gray value after filtering i.e.

$$POSP = \left[\frac{\text{Number of original pixels change their grayscale}}{\text{Number of non - noisy pixels}}\right] \times 100 \quad (18)$$

**Percentage of Noise Attenuated (PONA):** It may be defined as the number of pixels getting improved after being filtered.

$$PONA = \left[\frac{\text{Number of noisy pixels getting improved}}{\text{Total number of noisy pixels}}\right] \times 100 \quad (19)$$

This parameter reflects the capability of the impulse noise detector used prior to filtering. Small value of POSP shows the improved performance of the detector. High value of POSP leads to removal of original image properties and. edge jitters in restored images. The more the percentage the better will be the attenuation characteristic of the filter filtering. The proposed algorithm *NDB_RESTORE* will act accordingly.

---

*Algorithm: NDB_RESTORE*

Take 3X3 windows size for image.

Calculate the number of corrupted pixels in the windows.

repeat {

   if(n=0)

      pixels are unaltered.

   else if(n≤4)

      perform 3X3 median filtering.

   else if(5<n≤12)

      perform 5X5 median filtering.

   else if(n≥4)

   {

      perform 3X3 median filtering.

      if all the pixels in 3X3 windows is corrupted

         assume 5X5 windows size.

      else

         Replace the processed pixel by average of

         uncorrupted pixel.

} until next window.

Stop.

---

## V. IMPLEMENTATION

The proposed adaptive median/mean algorithm is applied to video sequences degraded by scratches, blotches, and impulses. Adaptive rood pattern search block matching algorithm [11] is used for motion estimation of the image sequences. Motion estimation and compensation techniques [11] are employed for tracking scratches on frames. Prediction and interpolation are used to estimate motion vectors for video de-noising. For fast motion prediction, commonly used technique is Block Matching (BM) motion estimator. The motion vector is obtained by minimizing a cost function measuring the mismatch between a block and each predictor candidate. The Motion Estimation (ME) gives motion vector of each pixel or block of pixels which is an essential tool for determining motion trajectories. Due to motion of objects in scene (i.e., corresponding regions in an image sequence), the same region does not occur in the same place in the previous frame as in current one. ARPS [11] algorithm makes use of the fact that the general motion in a frame is usually coherent, that is, if the macro blocks around the current macro block moved in a particular direction, then there is a high probability that the current macro block will also have a similar motion vector. ARPS algorithm uses the motion vector of the macro block to its immediate left to predict its own motion vector. The rood pattern search directly puts the search in an area where there is a high probability of finding a good matching block. The point that has the least weight becomes the origin for subsequent search steps, and the search pattern is changed to Small Diamond Search Pattern (SDSP). SDSP is repeated until least weighted point is found to be at the center of the SDSP. The main advantage of this algorithm over diamond search (DS) is that if the predicted motion vector is (0, 0), it does not waste computational time in carrying out Large Diamond Search Pattern (LDSP); it rather directly starts using SDSP.

The temporal median filter smoothes out sharp transitions in intensity at each pixel position; it not only de-noises the whole frame and removes blotches but also helps in stabilizing the illuminating fluctuations. Temporal median filtering removes the temporal noise in the form of small dots and streaks found in some videos. In this approach, dirt is viewed as a temporal impulse (single-frame incident) and hence treated by inter-frame processing by taking into account at least three consecutive frames.

## VI. RESULTS

The algorithm is tested with different types of degradations, namely, strip lines, drop lines, band missing, blotches, and impulse noise. The results are compared with those of general median filter, Lin's adaptive length median filter, Gonzalez adaptive length median filter and decision-based median filter.

The median filter and Lin's algorithm cause blur in the images and do not remove the degradations (Figures 6(c) and 6(d)). The Gonzalez adaptive algorithm removes the strip lines and drop lines but the edges are not preserved properly (Figures 6(d) and 7(d)) and this algorithm acts very poorly on the blotches and band noises. The proposed algorithm removes all these degradations more effectively with reduced blurring and edge preservation. The results of the removal of noise at different densities along with degradations.

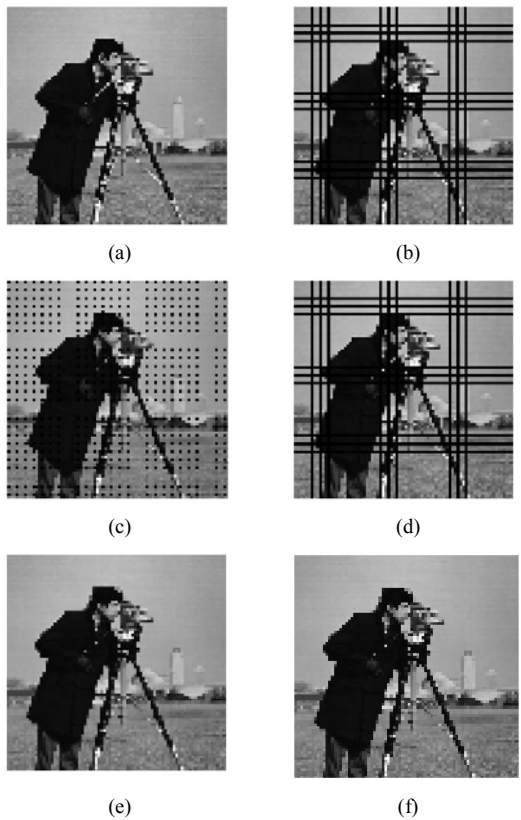

Fig. 6: Drop lines removal. (a) Original image. (b) Corrupted by drop lines. (c) Median filtered image. (d) Lin's adaptive length filter. (e) Gonzalez adaptive length filter. (f) Proposed algorithm.

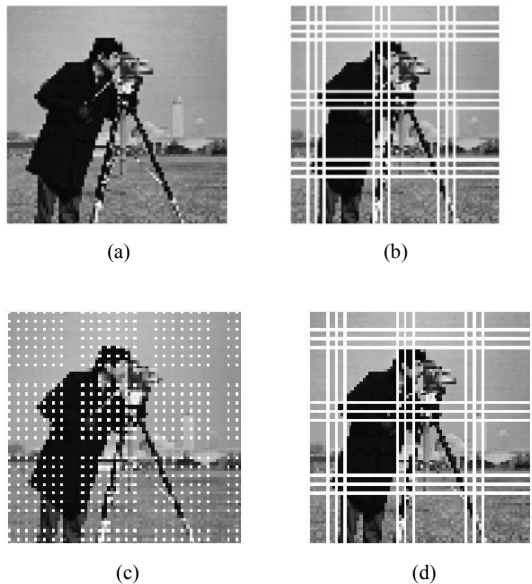

Fig. 7: Strip lines removal. (a) Original Image. (b) Corrupted by strip lines. (c) Median filtered image. (d) Lin's adaptive length filter. (e) Gonzalez adaptive length filter. (f) Proposed algorithm..

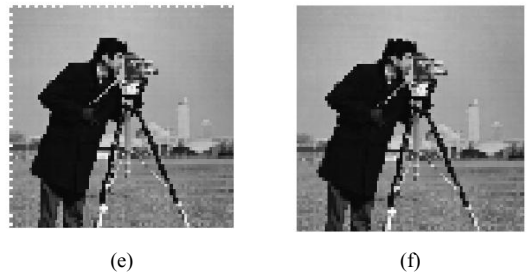

TABLE I
PSNR OF LENA AND GOLDHILL IMAGE CORRUPTED BY 20% OF IMPULSE NOISE AND THE PROPOSED ALGORITHM CORRUPTED BY 20% NOISE + DEGRADATIONS.

| Filter | Lena | Goldhill |
|---|---|---|
| SMF | 31.42 | 29.60 |
| CWMF [12] | 30.39 | 29.87 |
| DBMF [13] | 35.12 | 33.31 |
| TSMF [14] | 31.84 | 31.53 |
| ACWMF [8] | 36.54 | 34.42 |
| PF (noise + degradations) | 35.15 | 35.05 |

## VII. CONCLUSION

An adaptive length median/mean algorithm for removal of drops lines, strip lines, white bands, black bands, blotches, and impulses with minimum of blurring is developed. The performance is evaluated in terms of PSNR. The performance is compared with Lin's adaptive median filter, Gonzalez adaptive median filter, weighted median filter, decision-based median filter and adaptive center weighted median filter. The results show that the algorithm is more effective in the removal of drop lines, strip lines, white bands, black bands, and blotches along with impulse noise varying upto 70%. The advantage of the proposed algorithm is that a single algorithm with improved performance can replace several independent algorithms required for removal of different artifacts. Application of the proposed algorithm to black and color video sequences is also illustrated.

## ACKNOWLEDGMENT

The authors wish thanks to Dr. A. Jagadeesh, Director, RCET-Bhilai for his kind support. We would also especially grateful to Mr. Saurabh Rungta for providing necessary facilities to incorporate this research work.

## REFERENCES

[1] J. Astola and P. Kuosmanen, Fundamentals of Nonlinear Digital Filtering, CRC Press, New York, NY, USA, 1977.
[2] I. Pitas and A. N. Venetsanopoulos, Nonlinear Digital Filters: Principles and Applications, Kluwer Academic Publishers, Boston, Mass, USA, 1990.
[3] S. M. Shahrokhy, "Visual and statistical quality assessment and improvement of remotely sensed images," in Proceedings of the 20th Congress of the International Society for Photogrammetry and Remote Sensing (ISPRS '04), pp. 1–5, Istanbul, Turkey, July 2004.

[4] A. U. Silva and L. Corte-Real, "Removal of blotches and line scratches from film and video sequences using a digital restoration chain," in Proceedings of the IEEE-EURASIP Workshop on Nonlinear Signal and Image Processing (NSIP '99), pp. 826–829, Antalya, Turkey, June 1999.

[5] A. Kokaram, "Detection and removal of line scratches in degraded motion picture sequences," in Proceedings of the 8th European Signal Processing Conference (EUSIPCO '96), vol. 1, pp. 5–8, Trieste, Italy, September 1996.

[6] H.-M. Lin and A. N. Willson, Jr., "Median filters with adaptive length," IEEE Transactions on Circuits and Systems, vol. 35, no. 6, pp. 675–690, 1988.

[7] H. Hwang and R. A. Haddad, "Adaptive median filters: new algorithms and results," Transactions on Image Processing, vol. 4, no. 4, pp. 499–502, 1995.

[8] T. Chen and H. R. Wu, "Adaptive impulse detection using center-weighted median filters," IEEE Signal Processing Letters, vol. 8, no. 1, pp. 1–3, 2001.

[9] W. Luo, "An efficient detail-preserving approach for removing impulse noise in images," IEEE Signal Processing Letters, vol. 13, no. 7, pp. 413–416, 2006.

[10] K. S. Srinivasan and D. Ebenezer, "A new fast and efficient decision-based algorithm for removal of high-density impulse noises," IEEE Signal Processing Letters, vol. 14, no. 3, pp. 189–192, 2007.

[11] Y. Nie and K.-K. Ma, "Adaptive rood pattern search for fast block-matching motion estimation," IEEE Transactions on Image Processing, vol. 11, no. 12, pp. 1442–1449, 2002.

[12] S.-J. Ko and Y. H. Lee, "Center weighted median filters and their applications to image enhancement," IEEE Transactions on Circuits and Systems, vol. 38, no. 9, pp. 984–993, 1991.

[13] D. A. F. Florencio and R. W. Schafer, "Decision-based median filter using local signal statistics," in Visual Communications and Image Processing '94, vol. 2308 of Proceedings of SPIE, pp. 268–275, Chicago, Ill, USA, September 1994.

[14] T. Chen, K.-K. Ma, and L.-H. Chen, "Tri-state median filter for image denoising," IEEE Transactions on Image Processing, vol. 8, no. 12, pp. 1834–1838, 1999.

[15] E. Abreu, M. Lightstone, S. K. Mitra, and K. Arakawa, "A new efficient approach for the removal of impulse noise from highly corrupted images," IEEE Transactions on Image Processing, vol. 5, no. 6, pp. 1012–1025, 1996.

**Susanta K. Satpathy** (M'08) is working as a professor at Rungta College of Engineering and Technology,Bhilai, Chhattisgarh. He received the M. Tech degree in Computer Science and Engineering from NIT, Rourkela in 2003 and B.Tech. dgree in Computer Science and Engineering in 1995. He has published above 10 papers in various Journals and conferences. He is life time member of CSI and ISTE since 2008. He worked in various other engineering colleges for about 14 years. His area of reasearch includes Signal processing, image processing and information system and security.

**Sidhartha Panda** is working as a Professor at National Institute of Science and Technology (NIST), Berhampur, Orissa, India. He received the Ph.D. degree from Indian Institute of Technology, Roorkee, India in 2008, M.E. degree in Power Systems Engineering from UCE, Burla in 2001 and B.E. degree in Electrical Engineering in 1991. Earlier, he worked as an Associate Professor in KIIT Deemed University and also in various other engineering colleges for about 15 years. He has published above 50 papers in various International Journals and acting as a reviewer for some IEEE and Elsevier Journals. His biography has been included in Marquis' "Who's Who in the World" USA, for 2010 edition. His areas of research include power system transient stability, power system dynamic stability, FACTS, optimization techniques, model order reduction, distributed generation, image processing and wind energy.

**Kapil K. Nagwanshi** (M'09) was born in August 1978 in Chhindwara district, Madhya Pradesh, India. He is graduated from GG University, Bilaspur, a Central University of Chhattisgarh State, India, in Computer Science & Engineering in the year 2001 and later did his post graduation in Computer Technology & Application from Chhattisgarh Swami Vivekanand Technical University, Bhilai, India. He is life time member of International Association of Computer Science & Information Technology since 2009.Currently he is working as a Reader in RCET Bhilai, and he has published more than 28 research papers in reputed journals, national and international conferences. His research area includes, signal processing, and image processing, and information systems and security.

**Santanu K. Nayak** is working as Professor at Aditya Institute of Technology and Management (after taking lien from Berhampur University, Orissa). He completed his M.Tech. degree from Indian Institute of Science, Bangalore in Instrumentation in the year 1990 and joined as faculty member in the Department of Electronics(Berhampur University) in the year 1991. He received his Ph.D. degree in Electronics in the year 1996 from Berhampur University. He is member of IEEE, member and life member of different national professional bodies. He has published many papers in various national, international journals and conferences. He is also reviewer of EIS (Elsevier). His present area of research includes signal processing, image processing, fuzzy logic, instrumentation and embedded systems.

**C. Ardil** is with National Academy of Aviation, AZ1045, Baku, Azerbaijan, Bina, 25th km, NAA